\documentclass[letterpaper]{article}
\usepackage{proceed2e}
\usepackage[margin=1in]{geometry}

\usepackage{amsmath,amsthm}
\usepackage{euscript,microtype}

\usepackage[utf8]{inputenc} 
\usepackage[english]{babel}
\usepackage{csquotes}

\usepackage{amsmath, amssymb,bm, cases, mathtools, thmtools}
\usepackage{verbatim}
\usepackage{graphicx}\graphicspath{{figures/}}
\usepackage{multicol}
\usepackage{tabularx}
\usepackage[usenames,dvipsnames]{xcolor}
\usepackage{mathrsfs} 
\usepackage{caption}
\usepackage{algorithm}
\usepackage{algorithmicx}
\usepackage[noend]{algpseudocode}
\usepackage{textcomp}

\usepackage[%
    minnames=1,maxnames=99,maxcitenames=3,
    style=alphabetic, %
    doi=false,url=false,
    firstinits=true,
    hyperref,natbib,backend=biber]{biblatex}
\renewbibmacro{in:}{%
  \ifentrytype{article}{}{\printtext{\bibstring{in}\intitlepunct}}}
\renewcommand{\UrlFont}{\small\tt}

\bibliography{biblio}

\usepackage[colorlinks,citecolor=blue,urlcolor=darkgray,linkcolor=RawSienna]{hyperref}
\usepackage{hypernat}
\usepackage{datetime}

\DeclareMathAlphabet\EuRoman{U}{eur}{m}{n}
\SetMathAlphabet\EuRoman{bold}{U}{eur}{b}{n}

\usepackage[capitalize]{cleveref}

\crefname{lemma}{Lemma}{Lemmas}
\crefname{corollary}{Corollary}{Corollaries}
\crefname{theorem}{Theorem}{Theorems}

\makeatletter
\let\reftagform@=\tagform@
\def\tagform@#1{\maketag@@@{\ignorespaces\textcolor{gray}{(#1)}\unskip\@@italiccorr}}
\renewcommand{\eqref}[1]{\textup{\reftagform@{\ref{#1}}}}
\makeatother

\declaretheorem[style=plain,numberwithin=section,name=Theorem]{theorem}
\declaretheorem[style=plain,sibling=theorem,name=Lemma]{lemma}

\numberwithin{theorem}{section}

\def\[#1\]{\begin{align}#1\end{align}}
\def\*[#1\]{\begin{align*}#1\end{align*}}

\newcommand{\defas}{\vcentcolon=}  %

\newcommand{\Reals}{\mathbb{R}}

\newcommand{\Nats}{\mathbb{N}}

\newcommand{\NNReals}{\Reals_+}

\newcommand{\dee}{\mathrm{d}}

\DeclareMathOperator{\sign}{sign}

\DeclareMathOperator*{\newlim}{\mathrm{lim}\vphantom{\mathrm{infsup}}}
\DeclareMathOperator*{\newmin}{\mathrm{min}\vphantom{\mathrm{infsup}}}
\DeclareMathOperator*{\newmax}{\mathrm{max}\vphantom{\mathrm{infsup}}}
\DeclareMathOperator*{\newinf}{\mathrm{inf}\vphantom{\mathrm{infsup}}}
\DeclareMathOperator*{\newsup}{\mathrm{sup}\vphantom{\mathrm{infsup}}}
\renewcommand{\lim}{\newlim}
\renewcommand{\min}{\newmin}
\renewcommand{\max}{\newmax}
\renewcommand{\inf}{\newinf}
\renewcommand{\sup}{\newsup}

\renewcommand{\Pr}{\mathbb{P}}
\def\EE{\mathbb{E}}

\newcommand{\cF}{\mathcal F}

\newcommand{\Qsnn}{Q_{\mathrm{SNN}}}
\newcommand{\SigmaSNN}{\Sigma_{\mathrm{SNN}}}

\newcommand{\Sym}{\mathbb{S}}

\newcommand{\KLname}{\mathrm{KL}}
\newcommand{\HH}{\mathcal H}
\newcommand{\KL}[2]{\KLname(#1||#2)}

\newcommand{\BIGKLBOUND}[2]{\KLname^{-1}\biggl(#1 \biggl\lvert#2\biggr)}
\newcommand{\KLBOUND}[2]{\KLname^{-1}(#1|#2)}

\newcommand{\PACBayes}{PAC-Bayes}
\newcommand{\REB}{B_{\mathrm{RE}}}
\newcommand{\PosReals}{\Reals_{+}}

\newcommand{\NEWTON}{\mathrm{N}}

\newcommand{\diag}{\mathrm{diag}}

\newcommand{\Normal}{\mathcal N}
\newcommand{\trace}{\mathrm{tr}}
\newcommand{\Bernoulli}[1]{\mathcal B(#1)}

\newcommand{\EEE}[1]{\underset{#1}{\EE}}
\newcommand{\PPr}[1]{\underset{#1}{\Pr}}

\newcommand{\XX}{\Reals^{\idim}}
\newcommand{\YY}{\{-1,1\}}

\newcommand{\ce}{e}
\newcommand{\ece}{\hat \ce}

\newcommand{\loss}{\ell}

\newcommand{\surloss}{\breve\loss}
\newcommand{\surce}{\breve\ce}

\newcommand{\idim}{k}
\newcommand{\pdim}{d}

\newcommand{\datadist}{\mu}

\newcommand{\trainset}{S}

\newcommand{\e}{\mathrm{e}}

\newcommand{\cost}{\text{B}}

\newcommand{\wsgd}{w_{\text{SGD} }}
\newcommand{\wsnn}{w_{\text{SNN} }}

\newcommand{\hproduct}{\odot}

\usepackage{times}

\usepackage[explicit]{titlesec}
\usepackage{lipsum}

\titleformat{\section}
  {\large\bfseries}{\thesection}{1em}{\MakeUppercase{#1}}

\titleformat{\subsection}
  {\normalfont\bfseries}{\thesubsection}{1em}{\MakeUppercase{#1}}

\title{
Computing Nonvacuous Generalization Bounds for Deep (Stochastic) Neural Networks with 
Many More Parameters than Training Data
}

\author{} %

\author{ {\bf Gintare Karolina Dziugaite} \\ 
Department of Engineering \\
University of Cambridge
\And
{\bf Daniel M. Roy}  \\
Department of Statistical Sciences      \\
University of Toronto 
}

\newcommand{\LATER}[1]{\error}
\newcommand{\fLATER}[1]{\error}
\newcommand{\TBD}[1]{\error}
\newcommand{\fTBD}[1]{}
\newcommand{\PROBLEM}[1]{\error}
\newcommand{\fPROBLEM}[1]{\error}

\begin{document}

\maketitle

\begin{abstract}
One of the defining properties of deep learning is that models are chosen to have many more parameters than available training data.
In light of this capacity for overfitting, it is remarkable that simple algorithms like SGD reliably 
return solutions with low test error.  
One roadblock to explaining these phenomena in terms of implicit regularization, structural properties of the solution, and/or easiness of the data
is that many learning bounds are quantitatively vacuous when applied to networks learned by SGD 
in this ``deep learning'' regime. 
Logically, in order to explain generalization, we need nonvacuous bounds.
We return to an idea by Langford and Caruana (2001), who used PAC-Bayes bounds
to compute nonvacuous numerical bounds on generalization error for \emph{stochastic} two-layer two-hidden-unit neural networks via a sensitivity analysis.
By optimizing the PAC-Bayes bound directly, we are able to extend their approach and obtain 
nonvacuous generalization bounds 
for deep stochastic neural network classifiers with millions of parameters trained on only tens of thousands of examples.  We connect our findings to recent and old work on flat minima and MDL-based explanations of generalization.
\end{abstract}

\newcommand{\Rad}[1]{\mathcal R_{m}(#1)}
\newcommand{\cW}[1]{\mathcal W(#1)}
\newcommand{\textapprox}{\raisebox{0.25ex}{\texttildelow}}
\renewcommand{\UrlFont}{}

\section{Introduction}

By optimizing a PAC-Bayes bound, we show that it is possible 
to compute nonvacuous numerical bounds on the generalization error of 
deep \emph{stochastic} neural networks with millions of parameters, 
despite the training data sets being one or more orders of magnitude smaller than the number of parameters.
To our knowledge, these are the first explicit and nonvacuous numerical bounds 
computed for trained neural networks in the modern deep learning regime where the number of network parameters eclipses the number of training examples.

The bounds we compute are data dependent, 
incorporating millions of components optimized numerically to 
identify a large region in weight space with low average empirical error
around the solution obtained by stochastic gradient descent (SGD).
The data dependence is essential: 
indeed, 
the VC dimension of neural networks is typically bounded below by the number of parameters,
and so one needs as many training data as parameters before (uniform) PAC bounds are nonvacuous, 
i.e., before the generalization error falls below 1.
To put this in concrete terms, on MNIST, having even 72 hidden units in a fully connected first layer  
yields vacuous PAC bounds. 

Evidently, we are operating far from the worst case:
observed generalization cannot be explained in terms the regularizing effect of the size of the neural network alone.
This is an old observation, and one that attracted considerable theoretical attention 
 two decades ago: 
Bartlett \citep{bartlett1997valid,bartlett1998sample} 
showed that, 
in large (sigmoidal) neural networks,
when the learned weights are small in magnitude, 
the fat-shattering dimension is more important than the VC dimension for characterizing generalization.
In particular, Bartlett established classification error bounds in terms of the empirical margin and the fat-shattering dimension,
and then gave fat-shattering bounds for neural networks in terms of the \emph{magnitudes} of the weights and the depth of the network alone.
Improved norm-based bounds were obtained
using Rademacher and Gaussian complexity
by
\citet{bartlett2002rademacher,koltchinskii2002empirical}. 

These norm-based bounds are the foundation of our current understanding of neural network generalization.
It is widely accepted that these bounds %
explain observed generalization, at least ``qualitatively'' and/or when the
weights are explicitly regularized.
Indeed, recent work by \citet{Ney1412} puts forth the idea that SGD performs implicit norm-based regularization.
Somewhat surprisingly,
when we investigated state-of-the-art Rademacher bounds for ReLU networks,
the bounds were vacuous when applied to solutions obtained by SGD on real networks/datasets.
We discuss the details of this analysis in \cref{sec:pathnorm}.
While most theoreticians would assume these bounds were numerically loose to \emph{some} extent, 
they might be surprised to learn that the bounds do not logically establish generalization on their own.
It is worth highlighting that this observation does not necessarily rule out the existence of nonvacuous bounds under the same or similar hypotheses. This is an important avenue to investigate.

\subsection{Understanding SGD}

Our investigation was instigated by recent empirical work by 
\citefullauthor{Rethinking17}~\citep{Rethinking17},
who show that 
stochastic gradient descent (SGD),
applied to deep networks with millions of parameters,
is:\vspace*{-.6em}
\begin{enumerate}
\item able to achieve $\approx 0$ training error on CIFAR10 and IMAGENET and still generalize (i.e., test error remains small, despite the potential for overfitting);

\item 
\vspace*{-.2em}
still able to achieve $\approx 0$ training error even after the labels are \emph{randomized}, and does so with only a small factor of additional computational time.
\end{enumerate}
\vspace*{-.6em}
Taken together, these two observations demonstrate that 
these networks have a tremendous capacity to overfit
 and yet SGD does not abuse this capacity as it optimizes the surrogate loss, despite the lack of explicit regularization.

It is a major open problem to explain this phenomenon.
A natural approach would be to show that, under realistic hypotheses, 
SGD performs implicit regularization or tends to find solutions that possess some particular structural property
 that we already know to be connected to generalization.
However, in order to complete the logical connection, 
we need an associated error bound to be nonvacuous in the regime of model size / data size
where we hope to explain the phenomenon. 

This work establishes a potential candidate,
building 
off ideas by \citet{LangfordPHD} and \citet{LC02}:
On a binary class variant of MNIST, 
we find that SGD solutions are nearby to relatively large regions in weight space with low average empirical error.
We find this structure by optimizing a PAC-Bayes bound, starting at the SGD solution, 
obtaining a nonvacuous generalization bound for a stochastic neural network.
Across a variety of network architectures, our PAC-Bayes bounds on the test error are in the range 16--22\%.
These are far from nonvacuous but loose:
Chernoff bounds on the test error based on held-out data are consistently around 3\%.
Despite the gap, theoreticians aware of the numerical performance of generalization bounds
will likely be surprised that it is possible 
at all to obtain nonvacuous numerical bounds  
 for models with such large capacity trained on so few training examples. 
While we cannot entirely explain the magnitude of generalization, 
we can demonstrate nontrivial generalization.

Our approach was inspired by a line of work in physics 
by \citefullauthor{PhysRevLett.115.128101} \citep{PhysRevLett.115.128101} and the same authors with Borgs and Chayes \citep{BBCetal16}.
Based on theoretical results for discrete optimization linking computational efficiency to the existence of nonisolated solutions,
the authors propose a number of new algorithms for learning discrete neural networks by explicitly driving a local search towards nonisolated solutions. 
On the basis of Bayesian ideas, they posit that these solutions have good generalization properties.
In a recent work with 
Chaudhari, Choromanska, Soatto, and LeCun \citep{CCSL16},
they introduce local-entropy loss and EntropySGD,
extending these algorithmic ideas to modern deep learning architectures with continuous parametrizations,
and obtaining impressive empirical results.

In the continuous setting, nonisolated solutions correspond to ``flat minima''.
The existence and regularizing effects
of flat minima in the empirical error surface 
was recognized early on by researchers,
going back at work by
\citet{Hinton93} and \citet{Hochreiter97}.
\citeauthor{Hochreiter97} discuss sharp versus flat minima using the language of 
minimum description length (MDL; \citep{rissanen1983,grunwald2007minimum}). 
In short, describing weights in sharp minima requires high precision in order to not incur nontrivial excess error, 
whereas flat minimum can be described with lower precision.  A similar coding argument appears in \citep{Hinton93}.

\citefullauthor{Hochreiter97} propose an algorithm to find flat minima by minimizing the training error while maximizing the log volume of a connected region of the parameter space that yields similar classifiers with similarly good training error.
There are very close connections---at both the level of analysis and algorithms---with the work of \citet{CCSL16}
and close connections with the approach we take to compute nonvacuous error bounds 
by exploiting the local error surface. (We discuss more related work in \cref{sec:relwork}.)

Despite the promising underpinnings, 
the generalization theorems given by \citep{CCSL16} 
have admittedly unrealistic assumptions, 
and fall short of connecting local-entropy minimization
to observed generalization.

The goal of this work is to identify structure in the solutions obtained by SGD 
that provably implies small generalization error.
Computationally, it is much easier to demonstrate that a randomized classifier will generalize,
and so our 
results actually pertain to the generalization error of a \emph{stochastic} neural network, 
i.e., one whose weights/biases are drawn at random from some distribution on every forward evaluation of the network.
Under bounded loss, Fubini's theorem implies that we also obtain a bound on the expected error of a neural network whose weights have been randomly perturbed.
It would be interesting to achieve tighter control on the distribution of error or on the error of the mean neural network.

Returning to the goal of explaining SGD and generalization in deep learning more generally,
one could study whether the type of structure we exploit to obtain bounds 
necessarily arises from performing SGD under natural conditions.  
(We suspect one condition may be that the Bayes error rate is close to zero.)  
More ambitiously, perhaps the existence of the same structure can explain the success of SGD in practice.

\subsection{Approach}
\label{sec:approach}

Our working hypothesis is that SGD finds good solutions only if they are surrounded by a relatively large volume of solutions that are nearly as good. 
This hypothesis suggests that PAC-Bayes bounds may be fruitful: 
if SGD finds a solution contained in a large volume of equally good solutions, 
then the expected error rate of a classifier drawn at random from this volume should match that of the SGD solution.
The PAC-Bayes theorem \citep{PACBayes} bounds the expected error rate of a classifier chosen from a distribution $Q$ in terms of the Kullback--Liebler divergence from some a priori fixed distribution $P$, and so if the volume of equally good solutions is large, and not too far from the mass of $P$, we will obtain a nonvacuous bound.

Our approach will be to use optimization to find a broad distribution $Q$ over neural network parameters that minimizes the PAC-Bayes bound, in effect mapping out the volume of equally good solutions surrounding the SGD solution.
This idea is actually a modern take on an old idea by 
\citefullauthor{LC02}~\citep{LC02},
who apply PAC-Bayes bounds to small two-layer stochastic neural networks (with only $2$ hidden units)
that were trained on (relatively large, in comparison) data sets of several hundred labeled examples.

The basic idea can be traced back even further to work by \citefullauthor{Hinton93}~\citep{Hinton93},
who propose an algorithm for controlling overfitting in neural networks via the minimum description length principle.  
In particular, they minimize the sum of the empirical squared error and the KL divergence between a prior and posterior distribution on the weights.  
Their algorithm is applied to networks with 100's of inputs and 4 hidden units, trained on several hundred labeled examples.  
\citefullauthor{Hinton93} do not compute numerical generalization bounds to verify that MDL principles alone suffice to \emph{explain} the observed generalization.

Our algorithm more directly extends the work by \citeauthor{LC02},
who propose to construct a distribution $Q$ over neural networks by performing a sensitivity analysis on each parameter after training, searching for the largest deviation that does not increase the training error by more than, e.g., $1\%$. 
For $Q$, \citeauthor{LC02} choose a multivariate normal distribution over the network parameters, centered at the parameters of the trained neural network. 
The covariance matrix is diagonal, with the variance of each parameter chosen to be the estimated sensitivity, scaled by a global constant. 
(The global scale is chosen so that the training error of $Q$ is within, e.g., $1\%$ of that of the original trained network.) 
Their prior $P$ is also a multivariate normal, but with zero mean and covariance given by some scalar multiple of the identity matrix.  
By employing a union bound,
they allow themselves to choose the scalar multiple in a data-dependent fashion to optimize the PAC-Bayes bound.

The algorithm sketched by \citeauthor{LC02} does not scale to modern neural networks for several reasons, but one dominates:
in massively overparametrized networks, individual parameters often have negligible effect on the training classification error, and so it is not possible to estimate the \emph{relative}  sensitivity of large populations of neurons by studying the sensitivity of neurons in isolation.

Instead, we use stochastic gradient descent to directly optimize the PAC-Bayes bound on the error rate of a stochastic neural network.  
At each step, we update the network weights and their variances 
by taking a step along an unbiased estimate of the gradient of (an upper bound on) the PAC-Bayes bound.  In effect, the objective function is the sum of i) the empirical surrogate loss averaged over a random perturbation of the SGD solution, and ii) a generalization error bound that acts like a regularizer.

Having demonstrated that this simple approach can construct a witness to generalization,
it is worthwhile asking whether these ideas can be extended to the setting of local-entropic loss \citep{CCSL16}.
If we view the distribution that defines the local-entropic loss as defining a stochastic neural network,
can we use PAC-Bayes bounds to establish nonvacuous bounds on its generalization error?

\section{Preliminaries}

Much of our setup is identical to that of \citep{LC02}:
We are working in the batch supervised learning setting. 
Data points are elements $x \in \mathcal{X} \subseteq \XX$
with binary class labels $y \in \YY$.
Let $\trainset_m$ denote a training set of size $m$,
\begin{equation*}
\trainset_m = \{ (x_i, y_i) \}_{i=1,\ldots,m}, \text{ where  } (x_i, y_i ) \in \ (\mathcal{X}  \times \mathcal{Y}).
\end{equation*}
Let $\mathcal{M}$ denote the set of all probability measures on the data space $\XX \times \YY$. 
We will assume that the training examples are i.i.d. samples from some $\datadist \in \mathcal{M}$.

A parametric family of classifiers is a function $H : \Reals^{\pdim} \times \XX \to \YY$,
where $h_w \defas H(w,\cdot) : \XX \to \YY$ is the classifier indexed by the parameter $w \in \Reals^{\pdim}$.
The hypotheses space induced by $H$ is $\HH = \{h_w : w \in \Reals^{\pdim} \}$.
A randomized classifier is a distribution $Q$ on $\Reals^{\pdim}$.  Informally, we will speak of distributions on $\HH$ when we mean distributions on the underlying parametrization.

We are interested in the 0--1 loss $\loss :\Reals \times \YY \rightarrow \{0,1\} $ 
\begin{equation*}
\loss(\hat{y},y) = \mathbb{I} (\sign(\hat{y})=y).
\end{equation*}
We will also make use of the logistic loss $\surloss : \Reals \times \YY \rightarrow \Reals_{+}$
\begin{equation*}
\surloss (\hat{y}, y) = \frac{1}{\log(2)} \log \bigl(  1+ \exp(- \hat{y} y) \bigr),
\end{equation*}
which will serve as a convex surrogate (i.e., upper bound) to the 0--1 loss.

We define the following notions of error:
\begin{itemize}
\item $\displaystyle \ece(h,S_m) = \frac{1}{m} \sum_{i=1}^{m} \loss(h(x_i),y_i) $ 
empirical classification error of hypothesis $h$ for sample $S_m$;
\item $\displaystyle \surce (h,S_m) = \frac 1 m \sum_{i=1}^{m} \surloss (h(x_i), y_i) $ empirical (surrogate) error of a hypothesis $h$ on the training data set $S_m$.  We will use this for training purposes when we need our empirical loss to be differentiable;
\item $\displaystyle \ce_{\datadist}(h) = \EEE{S_m \sim \datadist^m} [\ece(h,S_m)]$   expected error for hypothesis $h$ under the data distribution $\datadist$ (we will often drop the subscript $\datadist$ and just write $\ce (h)$);
\item $\displaystyle \ece(Q, S_m) = \EEE{w \sim Q} [\ece(h_w, S_m)]$  expected empirical error under the randomized classifier $Q$ on $\HH$; 
\item $\displaystyle \ce(Q) =  \EEE{w \sim Q} [ \ce_{\datadist}(h_w) ]$ expected error for $Q$ on $\HH$.
\end{itemize}

\subsection{KL divergence}

Let $Q,P$ be probability measures defined on a common measurable space $\HH$, 
such that $Q$ is absolutely continuous with respect to $P$, 
and write $\frac{\dee Q}{\dee P} : \HH \to \NNReals \cup \{\infty\}$ for some Radon--Nikodym derivative of $Q$ with respect to $P$.
Then the Kullback--Liebler divergence (or relative entropy) of $P$ from $Q$ is defined to be
\begin{equation*}
\KL{Q}{P} 
\defas \int \log \frac{\dee Q}{\dee P} \,\dee Q.
\end{equation*}
We will mostly be concerned with KL divergences where $Q$ and $P$ are probability measures on Euclidean space, $\Reals^d$, 
absolutely continuous with respect to Lebesgue measure.  Let $q$ and $p$ denote the respective densities.
In this case, the definition of the KL divergence simplifies to
\begin{equation*}
\KL{Q}{P} 
= \int \log \frac{q(x)}{p(x)} q(x) \dee x.
\end{equation*}
Of particular interest to us is the KL divergence between multivariate normal distributions in $\Reals^d$.
Let $N_q = \Normal(\mu_q, \Sigma_q)$ be a multivariate normal with mean $\mu_q$ and covariance matrix $\Sigma_q$, let $N_p = \Normal(\mu_p, \Sigma_p)$,
and assume $\Sigma_q$ and $\Sigma_p$ are positive definite. Then $\KL{N_q}{N_p} $ is
\[
\label{normalKL} 
\begin{split}
\frac 1 2 
\biggl( \trace \left( \Sigma_p^{-1} \Sigma_q \right) - k &+ \left( \mu_p - \mu_q\right)^\top \Sigma_p^{-1} ( \mu_p - \mu_q ) \\
&+ \ln \left( \frac {\det \Sigma_p} {\det \Sigma_q}  \right)  
\biggr).
\end{split}
\]
For $p,q \in [0,1]$, we will abuse notation and define
\begin{align*}
\KL{q}{p} &\defas \KL{\Bernoulli{q}}{\Bernoulli{p}} \\
&= q \log \frac q p + (1-q) \log \frac {1-q}{1-p},
\end{align*}
where $\Bernoulli{p}$ denotes the Bernoulli distribution on $\{0,1\}$ with mean $p$.

\subsection{Inverting KL bounds}
\label{sec:inv}

In the following sections, we will encounter bounds on a quantity $p^* \in [0,1]$ of the form 
\begin{equation*}
\KL{q}{p^*} \le c
\end{equation*}
for some $q \in [0,1]$ and $c \ge 0$.
Thus, we are interested in
\begin{equation*}
\KLBOUND{q}{c} \defas \sup\, \{ p \in [0,1] \,:\, \KL{q}{p} \le c \}.
\end{equation*}
We are not aware of a simple formula for $\KLBOUND{q}{c}$,
although numerical approximations are readily obtained via Newton's method (\cref{sec:newtonapprox}).
For the purpose of gradient-based optimization,
we can use the well-known inequality, $2(q - p)^2 \le \KL{q}{p}$, to  obtain a simple upper bound 
\[\label{upperbound}
\KLBOUND{q}{c} \le q + \sqrt{c/2}.
\]
This bound is quantitatively loose near $q \approx 0$, because then
$\KLBOUND{q}{c} \approx c$ for $c \ll 1$, versus the upper bound of $\Theta(\sqrt{c})$.
On the other hand, when $c$ is large enough that $q + \sqrt{\frac {c}{2}} > 1$, the derivative of $\KLBOUND{q}{c}$ is zero, whereas the upper bound provides a useful derivative.

\subsection{Bounds}

We will employ three probabilistic bounds to control generalization error:  the union bound, a sample convergence bound derived from the Chernoff bound, and the PAC-Bayes bound due to \citet{PACBayes}. We state the union bound for completeness.

\begin{theorem}[union]\label{unionbound}
Let $E_1,E_2,\dots$ be events. Then $\Pr ( \bigcup_{n} E_n ) \le \sum_{n} \Pr (E_n)$.
\end{theorem}
 
Recall that $\Bernoulli{p}$ denotes the Bernoulli distribution on $\{0,1\}$ with mean $p \in [0,1]$. The following bound is derived from the KL formulation of the Chernoff bound:

\begin{theorem} [sample convergence {\citep{LC02}}]
\label{SampleConvBound}
For every $p,\delta \in (0,1)$ and $n \in \Nats$, with probability at least $1- \delta$ over $x \sim \Bernoulli{p}^{n}$,
$
  \KL{\textstyle n^{-1} \sum_{i=1}^{n} x_i}{p} \leq \frac{\log{\frac 2 \delta}}{n}. 
$
\end{theorem}

Finally, we present a variant of McAllester's PAC-Bayes bound due to \citet{LS01}.  (See also \citep{LangfordPHD}.)
\begin{theorem}[PAC-Bayes {\citep{PACBayes,LS01}}]
\label{pacbayes}
For every $\delta > 0$, $m \in \Nats$, distribution $\mu$ on $\XX \times \YY$, and distribution $P$ on $\HH$, with probability at least $1-\delta$ over $S_m \sim \mu^m$, for all distributions $Q$ on $\HH$,
\begin{equation*}
      \KL{\ece(Q,S_m)}{\ce(Q)} \le \frac {\KL{Q}{P} + \log \frac {m}{\delta} }{m-1}. 
\end{equation*}
\end{theorem}

The \PACBayes{} bound leads to the following learning algorithm \citep{PACBayes}:
\begin{enumerate}\setlength\itemsep{0em}
\item Fix a probability $\delta > 0$ and a distribution $P$ on $\HH$.
\item Collect an i.i.d.\ dataset $S_m$ of size $m$. 
\item Compute the optimal distribution $Q$ on $\HH$ that minimizes the error bound
\[\label{pacbayesbound}
\BIGKLBOUND{\ece(Q,S_m)}{\frac {\KL{Q}{P} + \log \frac {m}{\delta} }{m-1}}.
\]
\item Return the randomized classifier given by $Q$.
\end{enumerate}
In all but the simplest scenarios, making predictions according to the optimal $Q$ is intractable.  
However, we can attempt to approximate it.

\section{PAC-Bayes bound optimization}
\label{pacmethod}

Let $H$ be a parametric family of classifiers and write $h_w$ for $H(w,\cdot)$.
We will interpret $h_w$ as a neural network with (weight/bias) parameters $w \in \Reals^{\pdim}$, 
although the development below is more general.

Fix $\delta \in (0,1)$ and a distribution $P$ on $\Reals^{\pdim}$,
and let $S_m \sim \mu^m$ be $m$ i.i.d.\ training examples.
We aim to minimize the PAC-Bayes bound (\cref{pacbayesbound})
with respect to $Q$.

For $w \in \Reals^{\pdim}$ and $s \in \PosReals^{\pdim}$, 
let $\Normal_{w,s} = \Normal(w, \diag(s))$ denote the multivariate normal distribution with mean $w$ and diagonal covariance $\diag(s)$.
As our first simplifications, 
we replace the PAC-Bayes with the upper bound described by \cref{upperbound},
replace the empirical loss with its convex surrogate,
and restrict $Q$ to the family of multivariate normal distributions with diagonal covariance structure,
yielding the optimization problem
\begin{equation*}
\min_{w \in \Reals^{\pdim}, s \in \PosReals^{\pdim}} \ 
  \surce(\Normal_{w,s},S_m) 
 + \sqrt { 
        \frac {\KL{\Normal_{w,s}}{P} + \log \frac {m}{\delta} }{2(m-1)}
    }.
\end{equation*}

\subsection{The Prior}
In order to obtain a KL divergence in closed form, we choose $P$ to be multivariate normal.
Symmetry considerations would suggest that we choose $P = \Normal(0, \lambda I)$ for some $\lambda > 0$,
however there is no single good choice of $\lambda$. (We will also see that there are good reasons not to choose a zero mean, and so we will let $w_0$ denote the mean to be chosen a priori.)

In order to deal with the problem of choosing $\lambda$, we will follow \citet{LC02} and  
use a union-bound argument to choose $\lambda$ optimally from a discrete set, at the cost of a slight expansion to our generalization bound.
In particular, we will take $\lambda = c \exp\{ - j / b \}$ for some $j \in \Nats$ and fixed $b,c \ge 0$. 
(Hence, $b$ determines a level of precision and $c$ is an upper bound.)  
If the PAC-Bayes bound for each $j \in \Nats$ is designed to hold with probability at least 
$1-\frac 6 {\pi^2 j^2}$,
then, by the union bound (\cref{unionbound}), it will hold uniformly for all $j \in \Nats$ with probability at least  $1-\delta$, as desired.%
 During optimization, we will want to avoid discrete optimization, and so 
we will treat $\lambda$ as if it were a continuous variable. 
(We will then discretize $\lambda$ when we evaluate the PAC-Bayes bound after the fact.)
Solving for $j$, we have $j= b \log \frac {c}{\lambda}$, and so we will replace $j$ with this term during optimization.
Taking into account the choice of $P$ and the continuous approximation to the union bound, we have the following minimization problem:
\[\label{objective}
\min_{w \in \Reals^{\pdim}, s \in \PosReals^{\pdim}, \lambda \in (0,c)} \!\!\!\!\!
         \surce(\Normal_{w,s},S_m)  
        + \sqrt { \frac 1 2 \REB(w,s,\lambda;\delta) }                     
\]
where $\REB(w,s,\lambda;\delta)$ is
\[ \label{eq:REB}
             \frac {\KL{\Normal_{w,s}}{\Normal(w_0, \lambda I)}
                  \! + \! 2  \log( b \log \frac {c} {\lambda})
                    \! + \! \log \frac{\pi^2 m}{6 \delta} }
                  {m-1},
\]
and, using \cref{normalKL}, the KL term simplifies to
\begin{equation*}
\frac 1 2 
    \biggl(  \frac 1 \lambda \| s \|_1 
      \!-\! \pdim 
      + \frac 1 \lambda \| w - w_0  \|_2^2 
       + \! \pdim \log \lambda - 1_{\pdim} \cdot \log s
      \biggr).
\end{equation*}
We fix $\delta=0.025$, $b=100$, and $c=0.1$.

\subsection{Stochastic Gradient Descent}
We cannot optimize \cref{objective} directly because we cannot 
compute $\surce(\Normal_{w,s},S_m)$ or its gradients efficiently.
We can, however, compute the gradient of  
the unbiased estimate $\surce(h_{w+\xi \hproduct \sqrt{s}},S_m)$,  
where $\xi \sim \Normal_{0,I_d}$. 
We will use an i.i.d.\ copy of $\xi$ at each iteration.
We did not experiment using mini-batches during bound optimization.

\subsection{Final PAC-Bayes bound}
\label{sec:cbound}

While we treat $\lambda$ as a continuous parameter during optimization, 
the union bound requires that $\lambda$ be of the 
form $\lambda = c \exp\{ - j / b \}$, for some $j \in \Nats$. 
We therefore round $\lambda$ up or down, 
choosing that which delivers the best bound, 
as computed below.

According to the PAC-Bayes and union bound,
with probability $1-\delta$, 
uniformly over all $w \in \Reals^{\pdim}$, $s \in \PosReals^{\pdim}$, 
and $\lambda$ of the form $c \exp\{ - j / b \}$, for $j \in \Nats$,
the error rate of the randomized classifier $Q=\Normal_{w,s}$
is bounded by
\begin{equation*}
\KLBOUND{\ece(Q,S_m)}{\REB(w,s,\lambda;\delta)}.
\end{equation*}
We cannot compute this bound exactly because computing $\ece(Q,S_m)$ is intractable.  
However, we can obtain unbiased estimates and apply the sample convergence bound 
(\cref{SampleConvBound}). In particular, given $n$ i.i.d.\ samples $w_1,\dots,w_n$ from $Q$,
we produce the Monte Carlo approximation $\hat Q_n = \sum_{i=1}^n \delta_{w_i}$,
for which $\ece(\hat Q_n,S_m)$ is exactly computable, and 
obtain the bound 
\begin{equation*}
\begin{split}
\ece(Q,S_m) 
&\le \overline{\hat\ce_{n,\delta'}}(Q,S_m)  \\
& \defas \KLBOUND
             {\ece(\hat Q_n,S_m)}
             { n^{-1} \log 2/\delta' },
\end{split}
\end{equation*}
which holds with probability $1-\delta'$. 
By another application of the union bound, 
\[\label{finalbound}
\ce(Q) 
\le \KLBOUND
             {\overline{\hat\ce_{n,\delta'}}(Q,S_m)}
             {\REB(w,s,\lambda;\delta)},
\]
with probability $1-\delta-\delta'$. We use this bound in our reported results.

\begin{table*}[h]
\captionsetup{width=0.95\textwidth}
\vskip 0.25cm
\begin{center}
\begin{tabular}{l*{7}{c}}
Experiment              & T-600 & T-1200 & T-$300^2$ & T-$600^2$ & T-$1200^2$ &T-$600^3$ & R-600  \\
\hline
Train error            & 0.001 & 0.002  &  0.000 & 0.000 & 0.000 & 0.000 & 0.007   \\
Test error            & 0.018 & 0.018 &  0.015 & 0.016  & 0.015 &0.013 & 0.508  \\
SNN train error            & 0.028 & 0.027 & 0.027 & 0.028 & 0.029 & 0.027 & 0.112  \\
SNN test error            & 0.034 & 0.035  & 0.034 &0.033 & 0.035 & 0.032 & 0.503 \\
PAC-Bayes bound\!\!\!           &  0.161 &  0.179 & 0.170 & 0.186 & 0.223 & 0.201 & 1.352   \\
KL divergence        &  5144 & 5977 & 5791 & 6534 & 8558 &7861 & 201131 \\
$\#$ parameters         & 471k & 943k &  326k & 832k & 2384k & 1193k & 472k \\
VC dimension         & 26m &  56m &  26m  & 66m & 187m & 121m & 26m \\
\end{tabular}
\end{center}
\caption{Results for experiments on binary class variant of MNIST. 
SGD is either trained on (T) true labels or (R) random labels. 
The network architecture is expressed as $N^L$, indicating $L$ hidden layers with $N$ nodes each.
Errors are classification error. The reported VC dimension is the best known upper bound (in millions) for ReLU networks. The SNN error rates are tight upper bounds (see text for details). 
The PAC-Bayes bounds upper bound the test error with probability $0.965$.
}\label{resultstable} 
\end{table*}

\section{Experiments}

Starting from neural networks whose weights have been trained by 
SGD (with momentum) to achieve near-perfect accuracy on a binary class variant of MNIST,
we then optimize a PAC-Bayes bound on the error rate of stochastic neural network
whose weights are random perturbations of the weights learned by SGD.
We consider several different network architectures, varying both the depth and the width of the network. 

\subsection{Dataset}

We use the MNIST handwritten digits data set \citep{MNIST}
as provided in Tensorflow \citep{tensorflow}, 
where the dataset is split into the training set (55000 images) and test set (10000 images). (We do not use the validation set.)
Each MNIST image is black and white and 28-pixels square,
resulting in a network input dimension of $\idim = 784$. 
MNIST is usually treated as a multiclass classification problem.
In order to use standard PAC-Bayes bounds, we produce a binary classification problem
by mapping numbers $\{0,\dots,4\}$ to label $1$ and $\{5,\dots,9\}$ to label $-1$. 
In some experiments, we train on random labels, 
i.e., binary labels drawn independently and uniformly at random.

\subsection{Initial network training by SGD}

All experiments are performed on multilayer perceptrons, i.e., feed-forward neural networks with 2 or more layers, each layer fully connected to the previous and next layer.
We choose a standard initialization scheme for the weights and biases:
Weights are initialized randomly from a normal distribution (with mean zero and standard deviation $\sigma=0.04$) that is truncated to $[-2\sigma, 2\sigma]$.
Biases are initialized to a constant value of 0.1 for the first layer and 0 for the remaining layers. 
We let $w_0$ denote this random initialization of the weights (and biases). 

We use REctified Linear Unit (RELU) activations at every hidden node.
The last layer is linear.
In order to train the weights, we minimize the logistic loss by SGD with momentum 
(learning rate 0.01; momentum 0.9). 
SGD is run in mini-batches of size 100. 
These settings are similar to those in \citep{Rethinking17}.

On our binary class variant of MNIST, we train several neural network architectures of varying depth and width (see \cref{resultstable}). 
In each case, we train for a total of 20 epochs.
We also train a small network (with one 600-unit hidden layer) on \emph{random} labels,
in order to demonstrate the large capacity of the network. 
Obtaining $\approx$0 training error requires 120 epochs. 
See the first two rows of \cref{resultstable} for the train/test error rates.

\subsection{PAC-Bayes bound optimization}

Starting from weights $w$ learned by SGD,
we construct a stochastic neural network with   
a multivariate normal distribution $Q=\Normal_{w,s}$
over its weights 
with mean $w$ and covariance $\diag(s)$.
We initialize $s$ to $|w|$ and $|w|/10$
for true- and random-label experiments, respectively.

We optimize the PAC-Bayes bound (\cref{objective})
starting from an initial choice of $\e^{-6}$ for the prior variance $\lambda$
and the prior mean fixed at the random initialization $w_0$. 
(See \cref{sec:symmetries} for a discussion of this subtle but important innovation.)
We transform the constrained optimization over
 $w \in \Reals^{\pdim}$, $s \in \PosReals^{\pdim}$, and $\lambda \in (0,c)$,
into an unconstrained optimization over $w$,
$\frac 1 2 \log(s)$, and $\frac 1 2 \log(\lambda)$,
respectively.

We optimize the objective by gradient descent
with the RMSprop optimizer (with decay 0.9, as is typical).
We use the unbiased estimate of the gradient of the empirical surrogate error of the randomized classifier $Q=\Normal_{w,s}$.  
We set the learning rate to 0.001 for the first 150000 iterations, and then lower it to 0.0001 for the final 50000 iterations. 
For the random-label experiment, we optimize the bound with a smaller  0.0001 learning rate for 
500000 iterations. 
In both cases, the learning rate is tuned so that the objective decreases smoothly during learning.

\cref{algpac} is pseudo code for optimizing the PAC-Bayes bound. 
The code implements vanilla SGD,
although it can be easily modified to use an optimizer like RMSprop.

\subsection{Reported values}

Reported error rates correspond to classification error. 
Train and test error rates are empirical averages for networks learned by SGD. 
In light of 10000 test data points and the observed error rates, upper bounds via \cref{SampleConvBound} are only 0.005 higher.

Reported train and test error rates 
for the stochastic neural networks (abbreviated SNN) 
are upper bounds computed by an application of \cref{SampleConvBound} 
as described in \cref{sec:cbound} 
with $\delta' = 0.01$ and $n=150000$.
These numbers produce estimates within 0.001--0.002. 

The PAC-Bayes bound is computed as described in \cref{sec:cbound}.  
Each bound holds with probability 0.965 over the choice of the training set and the draws from the learned SNN $Q$. 
For the random-label experiment, we report $ \sqrt { \frac 1 2 \REB(w,s,\lambda;\delta)} $ from \cref{eq:REB}, since the PAC-Bayes bound is vacuous when this quantity is greater than 1.

Our VC-dimension bounds for ReLU networks are computed from a formula 
communicated to us by \citet{Bartlett17}. These bounds are in $O(L W \log W)$, where $L$ is the number of layers and $W$ is the total number of tunable parameters across layers. 

\section{Results}

See \cref{resultstable}. All SGD trained networks achieve perfect or near-perfect accuracy on the training data. 
On true labels, the SNN mean training error increases slightly as the weight distribution broadens to minimize the KL divergence. 
The SGD solution is close to mean of the SNN as measured with respect to the SNN covariance. (See \cref{sec:SNNclose} for a discussion.)
For the random-label experiment, the SNN mean training error rises above 10\%.
Ideally, it might have risen to nearly 50\%, while driving down the KL term to near zero.

The empirical test error of the SGD classifiers does not change much across the different architectures, 
despite the potential for overfitting.  
This phenomenon is well known, though still remarkable.
For the random-label experiment, the empirical test classification error of 0.508 represents lack of generalization, as expected. 
The same two patterns hold for the SNN test error too, with slightly higher error rates.

Remarkably,
the PAC-Bayes bounds 
do not grow much despite the networks becoming several times larger,
and all true label experiments have classification error bounded by 0.23. (This observation is consistent with \citep{Ney1412}.)
Since larger networks possess many more symmetries, the true PAC-Bayes bounds for our learned stochastic neural network classifier might be substantially smaller. (See \cref{sec:symmetries} for a discussion.)
While these bounds are several times larger than the test error estimated on held-out data (approximately, 0.03),
they demonstrate nontrivial generalization. 
The PAC-Bayes bound for the random-label experiment is vacuous.

The VC-dimension upper bounds indicate that data independent bounds will be vacuous by several orders of magnitude. 
Because the number of parameters exceeds the available training data, lower bounds imply that generalization cannot be explained in a data independent way.

\begin{algorithm}[t]
    \caption{PAC-Bayes bound optimization by SGD} 
    \label{algpac}
    \begin{algorithmic}[1] %
     \Require
     \Statex $w_0 \in \Reals^{\pdim}$ \Comment{Network parameters (random init.)}
     \Statex $w \in \Reals^{\pdim}$ \Comment{Network parameters (SGD solution)}
     \Statex $S_m$ \Comment{Training examples}
     \Statex $\delta \in (0,1)$
               \Comment{Confidence parameter}      
     \Statex $b \in \Nats, c \in (0,1)$ 
               \Comment{Precision and bound for $\lambda$} 
     \Statex $\tau \in (0,1), T$ \Comment{Learning rate; \# of iterations}
     \Ensure Optimal $w, s, \lambda$   \Comment{Weights, variances} %
        \Procedure{PAC-Bayes-SGD}{} %
     \State $\varsigma \gets  \mathrm{abs}(w)$ \Comment{where $s(\varsigma) = \e^{2 \varsigma}$}
     \State $\varrho \gets -3$ \Comment{where $\lambda(\varrho) = \e^{2\varrho}$}
     \State 
                  $\cost(w,s,\lambda,w') = \surce(h_{w'},S_m)+  \sqrt{\frac 1 2 \REB(w,s,\lambda)} $
            \For{$t\gets 1, T$} \Comment{Run SGD for T iterations.}
                 \State Sample $ \xi \sim \Normal(0,I_d)$
                 \State $w'(w,\varsigma) = w +  \xi \hproduct \sqrt{s(\varsigma)}$
                 \Statex \Comment{Gradient step}
                 \State $\displaystyle
                              \begin{bmatrix} w \\ \varsigma \\ \varrho \end{bmatrix}
                              \mathbin{\texttt{-=}} %
                                    \tau \! \begin{bmatrix} 
                                             \nabla_{w} B(w,s(\varsigma),\lambda(\varrho),w'(w,\varsigma)) 
                                          \\ \nabla_{\varsigma} B(w,s(\varsigma),\lambda(\varrho),w'(w,\varsigma))
                                          \\ \nabla_{\varrho} B(w,s(\varsigma),\lambda(\varrho),w'(w,\varsigma)) 
                                         \end{bmatrix}
                           $ 
            \EndFor
            \State \textbf{return} $w, s(\varsigma), \lambda(\varrho)$
        \EndProcedure
    \end{algorithmic}
\end{algorithm}

\section{Related work}
\label{sec:relwork}

As we mention in the introduction,
our approach scales the ideas in 
\citep{Hinton93} and \citep{LC02} to the modern deep learning regime where the networks have millions of parameters, but are trained on one or two orders of magnitude fewer training examples. 
The objective we optimize is an upper bound on the PAC-Bayes bound,  
which we know from the discussion in \cref{sec:inv}
will be very loose when the empirical classification error is approximately zero.
Indeed, in that case,
the PAC-Bayes bound is approximately
\[\label{altobj}
  \ece(\Normal_{w,s},S_m) 
 + \frac{\KL{\Normal_{w,s}}{P} + \log \frac {m}{\delta} }{(m-1)}.
\]
The objective optimized by \citefullauthor{Hinton93} is of the same essential form as this one,
except for the choice of squared error and different prior and posterior distributions.
We explored using \cref{altobj} as our objective with a surrogate loss, 
but it did not produce different results.

In the introduction we discuss the close connection of our work to several recent papers 
\citep{PhysRevLett.115.128101,BBCetal16,CCSL16} 
that study ``flat'' or nonisolated minima on the account of their generalization and/or algorithmic properties.

Based on 
theoretical results for k-SAT
 that efficient algorithms find nonisolated solutions, 
 \citet{BBCetal16} model efficient neural network learning algorithms 
as minimizers of a \emph{replicated} version of the empirical loss surface, 
which emphasizes nonisolated minima and deemphasizes isolated minima.
They then propose several algorithms for learning discrete neural networks using these ideas.

In follow-up work with
Chaudhari, Choromanska, Soatto, and LeCun \citep{CCSL16},
they translate these ideas into the setting of continuously parametrized neural networks.
They introduce an algorithm, called Entropy-SGD, which 
seeks out large regions of dense local minima: it maximizes the depth and flatness of the energy landscape. Their objective integrates both the energy of nearby parameters and the weighted distance to the parameters. In particular, rather than directly minimizing an error surface
$w \mapsto L(h_w,S_m)$,
they propose the following minimization problem over the so-called local-entropic loss:
\[\label{replicated}
\min_{w \in \Reals^{\pdim}} \quad \log \EEE{W \sim \Normal_{w,1/\gamma} }  [ C(\gamma) \exp\{-L (h_{W},S_m)\} ],
\]
where $\gamma > 0$ is a parameter and $C(\gamma)$ a constant.
In comparison, our algorithm can be interpreted as an optimization of the form
\[\label{approxprob}
\min_{w \in \Reals^p, s \in \PosReals^p}  \quad
      \EEE{W \sim \Normal_{w,s}}  [ L (h_{W},S_m) ]
        \,+\, R(w,s)
\]
where $R$ serves as a regularizer that accounts for the generalization error by, 
roughly speaking, 
trying to expand the axis-aligned ellipsoid 
$\{ x \in \Reals^{\pdim} : (w-x)^T \diag(s)^{-1} (w-x) = 1 \}$ 
and draw it closer to some point $w_0$ near the origin.
Comparing \cref{replicated,approxprob} highlights similarities and differences.
The local-entropic loss is sensitive to the volume of the regions containing good solutions. 
While the first term in our objective function looks similar, it does not, on its own, account for the volume of regions.  
This role is played by the second term, which prefers large regions (but also ones near the initialization $w_0$).
In our formulation, the first term is the empirical error of a stochastic neural network,
which is precisely the term whose generalization error we are trying to bound.
Entropy-SGD was not designed for the purpose of finding good stochastic neural networks, 
although it seems possible that having small local-entropic loss would lead to generalization for neural networks whose parameters are drawn from the local Gibbs distribution.
Another difference is that, in our formulation,
 the diagonal covariance of the multivariate normal perturbation is learned adaptively, 
and driven by the goal of minimizing error.  
The shape of the normal perturbation is not learned, although the region whose volume is being measured is determined by the error surface, and it seems likely that this volume will be larger than that spanned by a multivariate Gaussian chosen to lie entirely in a region with good loss.

\citet{CCSL16} give an informal characterization of the generalization properties of local-entropic loss in Bayesian terms 
by comparing the marginal likelihood of two Bayesian priors centered at a solution with small and large local-entropic loss.  
Informally, a Bayesian prior centered on an isolated solution will lead to small marginal likelihood in contrast to one centered in a wide valley. 
They give a formal result relying on the uniform stability of SGD \citep{HardtRS15} 
to show under some strong (and admittedly unrealistic) conditions that
Entropy-SGD generalizes better than SGD.  The key property is that the local-entropic loss surface is smoother than the original error surface.

Other authors have found evidence of the importance of ``flat'' minima:
Recent work by \citefullauthor{KMNST16}~\citep{KMNST16}  
finds that large-batch methods tend to converge to sharp / isolated minima and have worse generalization performance compared to mini-batch algorithms, which tend to converge to flat minima and have good generalization performance. 
The bulk of their paper is devoted to the problem of restoring good generalization behavior to batch algorithms.

Finally, our algorithm also bears resemblance to \emph{graduated optimization},
an approach toward non-convex optimization attributed to \citet{Blake:1987} 
whereby a sequence of increasingly fine-grained versions of an optimization problem are solved in succession. (See \citep{Hazan16} and references therein.)
In this context, \cref{replicated} is the result of a local smoothing operation acting on the  objective function $w \mapsto \surloss(h_w,S_M)$.
In graduate optimization, the effect of the local smoothing operation would be decreased over time, eventually disappearing.
In our formulation, the act of balancing the empirical loss and generalization error serve to drive the evolution of the local smoothing in an adaptive fashion.
Moreover, in the limit, the local smoothing does not vanish in our algorithm, as the volume spanned by the perturbations relates to the generalization error.
Our results suggest that SGD solutions live inside relatively large volumes, and so perhaps SGD can be understood in terms of graduated optimization.

\section{Conclusion and Future work} 

We obtain nonvacuous generalization bounds for deep neural networks with millions of parameters trained on 55000 MNIST examples.
These bounds are obtained by optimizing an objective derived from the PAC-Bayes bound, starting from the solution produced by SGD.
Despite the weights changing, 
the SGD solution remains well within the 1\% ellipsoidal quantile, i.e., the volume spanned by the stochastic neural network contains the original SGD solution.
(When labels are randomized, however, optimizing the PAC-Bayes bound causes the solution to shift considerably.) 

Our experiments look only at fully connected feed forward networks trained on a binary class variant of MNIST.  
It would be interesting to see if the results extend to multiclass classification, to other data sets, and to other types of architectures, especially convolutional ones. 

Our PAC-Bayes bound can be tightened in several ways. 
Highly dependent weights constrain the size of the axis-aligned ellipsoid representing the stochastic neural network.  We can potentially recognize small populations of highly dependent weights, and optimize their covariance parameters, rather than enforcing independence in the posterior. 

One might also consider replacing the multivariate normal posterior with a distribution that is more tuned to the loss surface.
One promising avenue is to follow the lines of \citet{CCSL16} and consider (local) Gibbs distributions. 
If the solutions obtained by minimizing the local-entropic loss are flatter than those obtained by  SGD, 
than we may be able to demonstrate quantitatively tighter bounds.

Finally, there is the hard work of understanding the generalization properties of SGD.  
In light of our work, 
it may be useful to start by asking whether SGD finds solutions in flat minima. 
Such solutions could then be lifted to stochastic neural networks with good generalization properties.
Going from stochastic networks back to deterministic ones may require additional structure.

\paragraph{Acknowledgments}

This research was carried out while the authors were visiting the Simons Institute for the Theory of Computing at UC Berkeley.
The authors would like to thank 
Peter Bartlett,
Shai Ben-David, 
Dylan Foster,
Matus Telgarsky,
and 
Ruth Urner
for helpful discussions.  
GKD is supported by an EPSRC studentship.  
DMR is supported by an
NSERC Discovery Grant, Connaught Award, and U.S. Air Force Office of Scientific
Research grant \#FA9550-15-1-0074.

\renewcommand*{\bibfont}{\small}
\printbibliography

\clearpage

\appendix

\section{Approximating $\KLBOUND{q}{c}$}
\label{sec:newtonapprox}

There is no simple formula for $\KLBOUND{q}{c}$, but we can approximate it via root-finding techniques. 
For all $q \in (0,1)$ and $c \ge 0$, define 
$h_{q,c}(p) = \KL{q}{p} - c$. Then $h'_{q,c}(p) = \frac{1-q}{1-p}-\frac{q}{p}$. Given a sufficiently good initial estimate $p_0$ of a root of $h_{q,c}(\cdot)$, we can obtain improved estimates of a root via Newton's method: 
\begin{equation*}
p_{n+1} = \NEWTON(p_{n};q,c)  \text{ where } \NEWTON(p;q,c) = p - \frac {h_{q,c}(c)}{h'_{q,c}(p)}.
\end{equation*}
This suggests the following approximation to $\KLBOUND{q}{c}$:
\begin{enumerate}
\item Let $\tilde b = q + \sqrt{\frac{c}{2}}$.
\item If $\tilde b \ge 1$, then return $1$.
\item Otherwise, return $\NEWTON^{k}(\tilde b)$, for some integer $k > 0$. 
\end{enumerate}
Our reported results use five steps of Newton's method. 

\section{Network symmetries}
\label{sec:symmetries}

In an ideal world, we would account for all the network symmetries when computing the KL divergence in the PAC-Bayes bound. 
However, it does not seem to be computationally feasible to account for the symmetries, as we discuss below.
Given this, it makes sense to try to break the symmetries somehow.  
Indeed, one consequence of randomly initializing a neural network's weights is that some symmetries are broken. 
If we do not expect SGD to reverse (many of) these symmetries, then the initial weight configuration, $w_0$, 
will be a better mean for the PAC-Bayes prior $P$ than the origin.  
In fact, breaking symmetries in this way lead to much better bounds than setting the means to zero.

\subsection{Bounds from mixtures}

Fix a neural network architecture $H : \Reals^{\pdim} \times \XX \to \YY$ and write $h_w$ for $H(w,\cdot)$.
It has long been appreciated that distinct parametrizations $w,w' \in \Reals^{\pdim}$ can lead to the same \emph{functions} $h_w = h_{w'}$,
and so the  set $\HH = \{ h_w : w \in \Reals^{\pdim} \}$ of classifiers defined by a neural network architecture
is a quotient space of $\Reals^{\pdim}$. %

For the purposes of understanding the generalization error of neural networks, 
we would ideally work directly with $\HH$.
Let $P,Q$ be a distributions on $\Reals^{\pdim}$, i.e., stochastic neural networks.
Then $P$ and $Q$ induce distributions on $\HH$, which we will denote by $\bar P$ and $\bar Q$, respectively.
For the purposes of the PAC-Bayes bound,  
it is the KL divergence $\KL{\bar Q}{\bar P}$ 
that upper bounds the performance of the stochastic neural network $Q$. 
In general, $\KL{\bar Q}{\bar P} \le \KL{Q}{P}$,
but it is difficult in practice to approximate the former because the quotient space is extremely complex.

One potential way to approach $\HH$ is to account for symmetries in the parameterization.
A \emph{network symmetry} is a map $\sigma: \Reals^{\pdim} \to \Reals^{\pdim}$ such that, for all $w \in \Reals^{\pdim}$,
we have $h_{w} = h_{\sigma (w)}$.  As an example of such a symmetry, in a fully connected network with identical activation functions at every unit, the function computed by the network is invariant to permuting the nodes with a hidden layer.
Let $\Sym$ be any finite set of symmetries possessed by the architecture.
For every distribution $Q$ on $\Reals^{\pdim}$ and network symmetry $\sigma$, 
we may define $Q_\sigma = Q \circ \sigma^{-1}$ to be the distribution over networks obtained by first sampling network parameters from $Q$ and then applying the map $\sigma$ to obtain a network that computes the same function.

Define $Q^{\Sym} = \frac 1 {|\Sym|} \sum_{\sigma \in \Sym} Q_{\sigma}$.  
Informally, $Q$ and $Q^{\Sym}$ are identical when viewed as distributions on functions, yet $Q^{\Sym}$ spreads its mass evenly over equivalent parametrizations.
In particular, for any data set $S$, we have $\ece(Q,S) = \ece(Q^{\Sym},S)$.
We call $Q^{\Sym}$ a symmetrized version of $Q$. 
The following lemma states that symmetrized versions always have smaller KL divergence with respect to distributions that are invariant to symmetrization:
Before stating the lemma, recall that the 
differential entropy of
an absolutely continuous distribution $Q$ on $\Reals^{\pdim}$ with density $q$ is
$\int q(x) \log q(x) \dee x \in \Reals \cup \{-\infty, \infty \}$.
\begin{lemma}
Let $\Sym$ be a finite set of network symmetries,
let $P$ be an absolutely continuous distribution such that $P=P_{\sigma}$ for all $\sigma \in \Sym$, and
define $Q^{\Sym}$ as above for some arbitrary absolutely continuous distribution $Q$ on $\Reals^{\pdim}$ with finite differential entropy.
Then $\KL{Q^{\Sym}}{P} = \KL{Q}{P} - \KL{Q}{Q^{\Sym}} \le \KL{Q}{P}$.
\end{lemma}
The above lemma can be generalized to distributions over (potentially infinite) sets of network symmetries.

It follows from this lemma that one can do no worse by accounting for symmetries using mixtures, provided that one is comparing to a distribution $P$ that is invariant to those symmetries. 
In light of the PAC-Bayes theorem, this means that a generalization bound based upon a KL divergence that does not account for symmetries  can likely be improved. 
However, for a finite set $\Sym$ of symmetries, it is easy to show that the improvement is bounded by $\log |\Sym|$,
which suggests that, in order to obtain appreciable improvements in a numerical bound, 
one would need to account for an exponential number of symmetries.
Unfortunately, exploiting this many symmetries seems intractable.  It is hard to obtain useful lower bounds 
to $\KL{Q}{Q^{\Sym}}$, while upper bounds from Jensen's inequality lead to negative (hence vacuous) lower bounds on $\KL{Q^{\Sym}}{P}$.

In this work, we therefore take a different approach to dealing with symmetries.
Neural networks are randomly initialized in order to \emph{break} symmetries.
Combined with the idea that the learned parameters will reflect these broken symmetries,
we choose our prior $P$ to be located at the random initialization, rather than at zero.

\section{Comparing weights before and after PAC-Bayes optimization}
\label{sec:SNNclose}

In the course of optimizing the PAC-Bayes bound,
we allow the mean $w$ to deviate from the SGD solution $\wsgd$ that serves as the starting point. 
This is necessary to obtain bounds as tight as those that we computed.
Do the weights change much during optimization of the bound? How would we measure this change?

To answer these questions, we calculated the p-value of the SGD solution
under the distribution of the stochastic neural network.

Let $\Qsnn$ denote the distribution obtained by optimizing the PAC-Bayes bound,
write $\wsnn$ and $\SigmaSNN$ for its mean and covariance,
and let $\| w \|_{\SigmaSNN} = w^T \SigmaSNN^{-1} w$ denote the induced norm.
Using 10000 samples, we estimated
\begin{equation*}
\PPr{w \sim Q_{\text{SNN}}}
     \Bigl (  
            \| w - \wsnn \|_{\SigmaSNN} 
            <  \| \wsgd - \wsnn \|_{\SigmaSNN} 
     \Bigr  ).
\end{equation*}
The estimate was $0$ for all true label experiments, 
i.e., $\wsgd$ is less extreme of a perturbation of $\wsnn$ than 
a typical perturbation.  
For the random-label experiments, $\wsnn$ and $\wsgd$ differ significantly,
which is consistent with the bound being optimized in the face of random labels.

\section{Evaluating Rademacher error bounds}
\label{sec:pathnorm}

Fix a class $\cF$ of measurable functions from $\Reals^D$ to $\Reals$
and let $\Rad{\cF}$ denote the Rademacher complexity of $\cF$ associated with $m$ i.i.d.\ samples.
For $h \in \cF$, we will obtain binary classifications (and measure error and empirical error) by computing the sign of its output, i.e., by thresholding.
The following error bound is a straightforward adaptation of
\citep[Thm.~7]{bartlett2002rademacher},
which is itself an adaptation 
of \citep[Thm.~2]{koltchinskii2002empirical}.

\begin{theorem}\label{radbound2}
For every $L > 0$, with probability at least $1-\delta$ over the choice of $S_m \sim \datadist^m$,
for all $h \in \cF$,
\[
\ce(h) 
&\le 
\ece(h,S_m,L) + 2 L \Rad{\cF} + \sqrt{\frac{\log (\frac 2 \delta)}{2 m}},
\]
where
\[
\ece(h,S_m,L) = \frac 1  m \sum_{i=1}^{m} \max( \min(1 - L y_i h(x_i),1), 0) . \nonumber
\]
\end{theorem}

In order to compute these bounds, we must compute (bounds on) the Rademacher complexity of appropriate function classes.
To that end, we will use results by \citet{Ney1503} for ReLU networks (i.e., multilayer perceptrons with ReLU activations).

Let $w$ be the weights of a ReLU network 
and let $w^{(k)}_{i,j}$ denote the weight associated with the edge from neuron $i$ in layer $k-1$ to neuron $j$ in layer $k$. 
\citet{Ney1503} define the $\ell_1$ path norm 
\[
\phi_1 (w) = \sum_{j} \biggl [ |w^{(2)}_{j,1} | \sum_{i} |w^{(1)}_{i,j}| \biggr ],
\]
stated here in the special case of a 2-layer network with 1 output neuron.
For any number of layers, the path norm can be computed easily in a forward pass, 
requiring only a matrix--vector product at each layer.

\citeauthor{Ney1503} also provide the follow Rademacher bound in terms of the path norm:

\begin{theorem}[{\citep[Cor.~7]{Ney1503}}] \label{radbound}
Given $m$ datapoints $x_1,\dots,x_m \in \Reals^D$,
the Rademacher complexity of the class of 
depth-$d$ ReLU networks,
whose $\ell_1$ path norms are bounded by $\phi$,
is no greater than
\[
2^d 
\phi 
\sqrt { \frac { \log(2 D)  } {m} }
\max_i \| x_i \|_{\infty}
.
\]
\end{theorem}

Let $w^{(k)}_j$ for the $j$th column of $w^{(j)}$, i.e., the vector of weights for edges from layer $k-1$ to neuron $j$ in layer $k$.
The $\ell_1$ path norm is closely related to the norm
\[
\gamma_{1,\infty}(w) = \prod_{i=1}^d  \max_j \, \bigl \| w^{(k)}_j \bigr \|_1. \nonumber
\]
If the upper bound $\phi$ appearing in the bound of \cref{radbound} is instead taken to be a bound on $\gamma_{1,\infty}(w)$,
then one essentially obtains the Gaussian complexity bounds for neural networks established by
\citet{bartlett2002rademacher,koltchinskii2002empirical}.
However, their bounds 
apply only to networks with bounded activation functions, ruling out ReLU networks.

Regardless, the path-norm bound is tighter for ReLU networks.
In order to establish the connection, 
let $\cW{w}$ denote the set of all weights $w'$ obtained from redistributing the weights $w$ across layers, i.e., by multiplying the weights $w^{(k-1)}$  in a layer by a constant $c>0$ and multiplying the weights in the subsequent layer $w^{(k)}$ by $c^{-1}$. Note that the function computed by a ReLU network is invariant to this transformation. 
This is the key insight of \citeauthor{Ney1503}.
Obviously, $\phi_1(w) = \phi_1(w')$ for all $w' \in \cW{w}$.
\citeauthor{Ney1503} show that 
$\phi_1(w) = \inf_{w' \in \cW{w}} \gamma_{1,\infty}(w')$,
and so the path norm better captures the complexity of a ReLU network.

In our experiments,
we will compute the 
bound obtained by combining \cref{radbound,radbound2}.

Note that the constant $L$ in \cref{radbound2} must be chosen independently of the data $S_m$.  
As in the original result \citep[Thm.~2]{koltchinskii2002empirical},
one can use a union bound to allow oneself to choose $L$ based on the data in order to minimize the bound. 
Even though the effect of this change is usually (relatively) small, 
its magnitude depends on the particular weight function employed in the union bound.
Instead, we will apply the bound with an optimized $L$, yielding an optimistic bound (formally, a lower bound on any upper bound obtained from a union bound). 
We optimize $L$ over a grid of values, and handle the vacuous edge cases analytically.
Nevertheless, we will see even the resulting (optimistic) bound is vacuous.

\subsection{Experiment details}

\begin{figure*}[t]
  \centering
    \includegraphics[width=.9\linewidth]{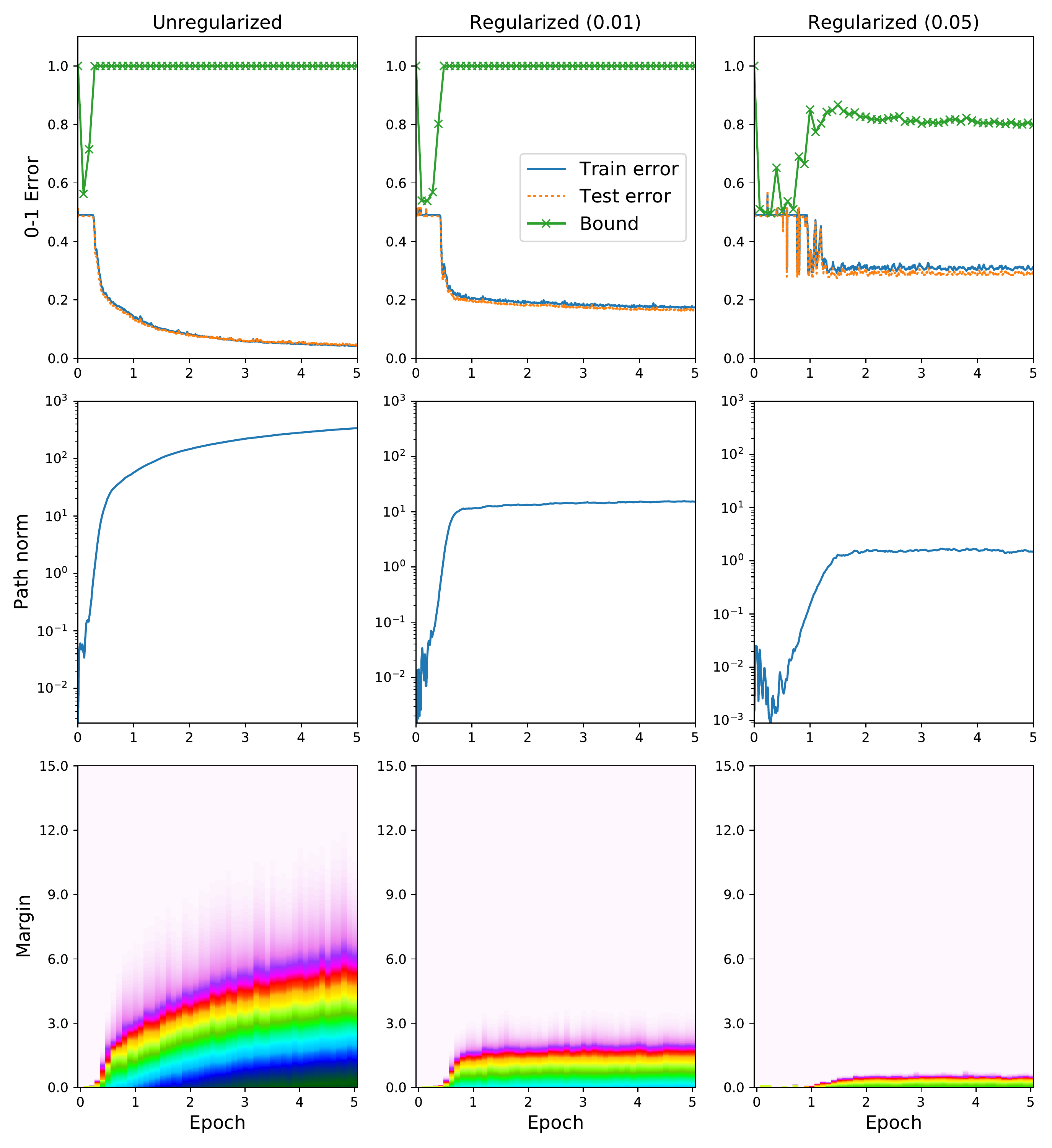} 

    \vspace*{-6.145cm} \hspace*{15.3cm}
    \includegraphics[height=5.92cm]{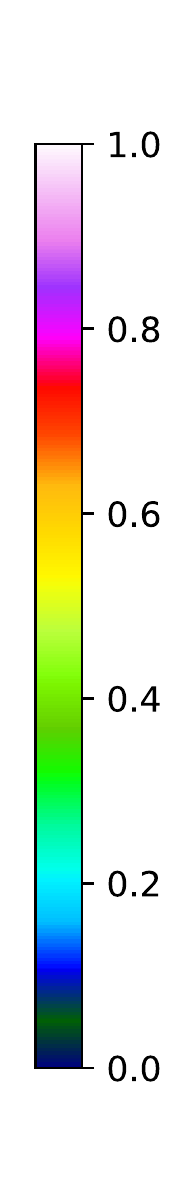} 

  \caption{Unregularized ({\bf left column}) 
  and 
  path-norm regularized ({\bf center and right columns} with regularization parameter specified in parenthesis) 
  optimization of two-layer 600-hidden-unit ReLU network by SGD for 5 epochs. (We ran 20 epochs and found no new patterns. Plots for longer experiment obscured the initial behavior.)
  ({\bf top row})
  Training error, testing error, and error bounds versus (iterations measured in) epochs.
   Without regularization, the bounds are immediately vacuous once the network performance deviates from chance, and this remains true under regularization unless the explicit regularization is very strong. In this case, the bound is nonvacuous, but trivial in the sense that the error rate of guessing is 50\%. Note that training/testing error is also very large in this case.
  ({\bf center row})
   Log plot of path norm versus epochs. Without regularization, the path norm diverges quickly.
  ({\bf bottom row}) 
  Empirical margin distributions versus epochs. The margin that attains a fixed average loss is growing, but not rapidly enough 
  to counteract the rapidly increasing path norm.
  }
\label{pathbnd}
\end{figure*}

We use SGD to train a two-layer 600-hidden-unit ReLU network on the same binary class variant of MNIST used to evaluate our PAC-Bayes bounds. 
We set the global learning rate to 0.005. 
As in our PAC-Bayes experiments, we optimize the average logistic loss during training.
The random initializations commonly used for ReLU networks lead to initial path norms that produce vacuous error bounds.
In order to visualize the behavior of the path-norm bound under SGD, we reduce
the standard deviation of the truncated-normal initialization from 0.04 to 0.0001. 
As before, we use mini-batches of 100 training examples, 
yielding 550 iterations per epoch.

For comparison, we also train the same network architecture while explicitly regularizing the path norm. 
(\citet{Ney1506} propose training neural networks via steepest descent with respect to the path norm.  We leave this comparison to future work.)

\subsection{Results}

When the network is trained by optimizing the logistic cost function without regularization, 
the error bound becomes vacuous within a fraction of a single epoch.  This occurs before the training error dips appreciable below chance.
The bound's behavior is due to the path norm diverging.  
While the level sets $\ece(h,S,\cdot)^{-1}$ of the empirical margin distribution are growing, they are not growing fast enough to counteract the growth of the path norm. (See the left column of \cref{pathbnd}.)

When the network is trained with explicit path-norm regularization, we obtain vacuous error bounds, unless we apply excessive amounts of regularization.
We report results when the regularization parameter is 0.01 and 0.05. 
Both settings are clearly too large, as evidenced by the training error converging to \textapprox{}20\% and \textapprox{}30\%, respectively.
 A cursory study of overall $\ell_1$ and $\ell_2$ regularization produced qualitatively similar results. Further study is necessary.

\end{document}